\documentclass[runningheads]{llncs}
\usepackage{graphicx}
\usepackage{amsmath}
\begin{document}
%
%\title{Physics-informed Feature Selection for Log-based Anomaly Detection and Prediction}
\title{Feature Selection for Fault Detection and Prediction based on Event Log Analysis}
\titlerunning{Feature Selection for Fault Detection and Prediction}

\author{Zhong Li\inst{1}\orcidID{
0000-0003-1124-5778} \and
Matthijs van Leeuwen \inst{1}\orcidID{0000-0002-0510-3549
}}
\authorrunning{Li and van Leeuwen}
\institute{Leiden Institute of Advanced Computer Science, Leiden University, Leiden, The Netherlands \email{\{z.li,
m.van.leeuwen\}@liacs.leidenuniv.nl}}
\maketitle              % typeset the header of the contribution
\begin{abstract}
 Event logs are widely used for anomaly detection and prediction in complex systems. Existing log-based anomaly detection methods usually consist of four main steps: log collection, log parsing, feature extraction, and anomaly detection, wherein the feature extraction step extracts useful features for anomaly detection by counting log events. For a complex system, such as a lithography machine consisting of a large number of subsystems, its log may contain thousands of different events, resulting in abounding extracted features. However, when anomaly detection is performed at the subsystem level, analyzing all features becomes expensive and unnecessary. To mitigate this problem, we develop a feature selection method for log-based anomaly detection and prediction, largely improving the effectiveness and efficiency. (This is a work-in-progress paper.) \footnote{This is a work-in-progress paper that was accepted by the 
 AI for Manufacturing Workshop at ECMLPKDD 2022, with oral presentation \& poster.}

\end{abstract}
\section{Introduction}
A lithography machine is a complex structural equipment used to manufacture chips. Typically, it consists of the following main subsystems: the light source subsystem, the objective lens subsystem, the table subsystem, the mask table
subsystem, the mask transfer subsystem, the wafer transfer subsystem, and the exposure subsystem \cite{zhang2021introduction}. Particularly, the wafer transfer subsystem serves to transfer silicon wafers between the track and wafer stage, having a great impact on the precision of chip fabrication. A wafer transfer subsystem usually contains two robots, namely a load robot and an unload robot. When the lithography machine goes into production, these two robots may encounter some faults. We assume there are $L+M$ types of faults, viz. $GF_{1}$,$GF_{2}$,...,$GF_{L}$ and $SF_{1}$,$SF_{2}$,...,$SF_{M}$. Specifically, $GF_{1}$,$GF_{2}$,...,$GF_{L}$ represent faults that occur gradually and thus can be detected in an early stage (e.g., they are predictable). In contrast, $SF_{1}$,$SF_{2}$,...,$SF_{M}$ denote faults that generally occur suddenly and are often hard to predict.

To minimize machine downtime and thus maximize productivity, the possible faults of load and unload robots should be detected and predicted (if possible) in an automated way. To this end, the wafer transfer subsystem usually uses sensors to measure the position of the two robots in real time, collecting time series data that can be used for data-driven fault detection and prediction. However, due to the limited information contained in sensor data, it is challenging to detect all possible faults based on time series data alone. Meanwhile, as shown in Table \ref{tab:event}, a lithography machine has an information system that records all triggered events in the form of logs when the machine is working. Since the different subsystems in a lithography machine are interconnected, a fault incurred in a subsystem (e.g. the wafer transfer system) may trigger events not only in that subsystem, but also in other subsystems. Besides, the components in the same subsystem are usually closely interconnected. Therefore, the fault of one component is very likely to cause faults of other components. Therefore, the event logs contain important information for fault detection and prediction. Traditional log-based anomaly detection methods can be used to detect such faults \cite{yadav2020survey}. 
 
 Due to the complexity of the lithography machine, there can be thousands of unique log events, resulting in millions of log events in a relatively short working time of the machine. However, when attempting to detect faults of certain components in a specific subsystem (e.g., the load and unload robots in the wafer transfer subsystem), many of these log events are irrelevant or abundant. Hence, a direct application of existing log-based anomaly detection methods on all log events can be computationally prohibitive and may also produce misleading detection results due to the inclusion of irrelevant log events. To mitigate this problem, we regard each log event as a feature and develop a feature selection method that aims to select relevant features for log-based fault detection and prediction. 
 
 In brief, our method consists of three main modules, namely \textit{Log Event Vectorization}, \textit{Selection of Relevant Features} and \textit{Removal of Redundant Features}. Specifically, the \textit{Log Event Vectorization} module aims at converting  unstructured log events into time series data; the \textit{Selection of Relevant Features} module attempts to select relevant features for fault detection and prediction by using the variables measured by sensors as target; and the \textit{Removal of Redundant Features} module focuses on eliminating redundant features to further reduce the number of selected features.

\section{Related Work}
Existing log-based anomaly detection methods usually consist of four main steps: \textit{Log Collection}, \textit{Log Parsing}, \textit{Feature Extraction} and \textit{Anomaly Detection} \cite{he2016experience}. First, the \textit{Log Collection} step is responsible for recording triggered events in the form of logs. A record is called an \textit{log message}, which usually contains the date and time of occurrence and the detailed description of event. More concretely, detailed descriptions are typically presented in predefined templates, and may also include parameters. Second, the \textit{Log Parsing} step aims at converting each log
message into a specific \textit{log event} template \cite{zhu2019tools}. Usually, a \textit{log event} corresponds to a unique template. Third, based on derived log events, the \textit{Feature Extraction} step attempts to convert each \textit{log sequence} into a \textit{log count vector}. Specifically, a \textit{log sequence} is composed of multiple \textit{log events}. In general, a \textit{log count vector} is a vector with each entry indicating the number of times that the corresponding \textit{log event} was triggered. Note that the entries of the \textit{log count vector} can be computed in other refined way \cite{guo2021logbert}. Finally, the \textit{Anomaly Detection} step performs anomaly detection based on extracted \textit{log count vectors}.

Since our work centers around feature selection (e.g., log event selection)  for fault detection and prediction, we will only consider the \textit{Feature Extraction} and \textit{Anomaly Detection} steps. Due to the novelty of the faced problems, we are not aware of any existing publications that are closely related to our work.

\section{Method}

\subsection{Terminology and Problem Formulation}

\begin{table}[bt]
\centering
\caption{An example event log. Note that all values are fictional.}\label{tab:event}
\begin{tabular}{|l|l|l|l|l|}
\hline
$\mathbf{Machine}$ &  $\mathbf{Code}$ & $\mathbf{Severity}$ & $\mathbf{Detail}$ & $\mathbf{DateTime}$\\
\hline
1 & AA-BBBB & Low & description & 2020-01-01 00:00:01\\
1 & CC-DDDD & Medium & description & 2020-01-01 00:00:01\\
1 & AA-BBBB & Low & description & 2020-01-01 00:01:00\\
1 & AA-BBBB & Low & description & 2020-01-01 00:02:03\\
1 & EE-FFFF & High & description & 2020-01-01 00:05:00\\
...&...&...&...&...\\
\hline
\end{tabular}
\end{table}

We assume that we access to two types of data. First, as shown in Table \ref{tab:event}, we assume that the log data in a lithography machine, denoted by $\mathbf{X}$, has been collected and well parsed. Without loss of generality, we suppose there is a $\mathbf{Code}$  as the unique identifier for each \textit{log event}, a $\mathbf{Severity}$  roughly indicating the severity level of triggered \textit{log event}, a $\mathbf{Detail}$  describing the detail of each \textit{log message} that is an instantiation of a \textit{log event} using a predefined template, and a $\mathbf{DateTime}$ containing the corresponding date and time. Hereinafter, we also call each \textit{log event} a \textit{log feature}. In addition, we may have log data for multiple lithography machines, and we use $\mathbf{Machine}$ to represent the corresponding name of the machine. 

Ideally, applying an existing log-based anomaly detection method on $\mathbf{X}$ can detect most faults related to load and unload robots. However, due to the large number of \textit{log features}, it is computationally prohibitive to directly use existing anomaly detection methods. Furthermore, the presence of irrelevant \textit{log features} may significantly degrade detection performance and even lead to misleading detection results.
 
 Second, we also assume the availability of sensor data, denoted by  $\mathbf{Y}$, that measures the positions of robots. As shown in Table \ref{tab:sensor}, there are measurements and corresponding timestamps from $K$ different positions for the load robot and the unload robot, respectively. By using the \textit{Log Event Vectorization} module in our proposed method (as will be explained in the sequel), $\mathbf{Y}$ can be rewritten as $(\mathbf{LP_{1}},...,\mathbf{LP_{K}},\mathbf{UP_{1}},...,\mathbf{UP_{K}})$. For $k \in \{1,...,K\}$, $\mathbf{LP_{k}} = \{(Value_{t},DateTime_{t})\}_{t=1}^{T}$ denotes the corresponding time series of load robot from position $k$ and $\mathbf{UP_{k}} = \{(Value_{t},DateTime_{t})\}_{t=1}^{T}$ denotes the corresponding time series of unload robot from position $k$, respectively.
 
 By applying an appropriate time series anomaly detection method on $(\mathbf{LP_{1}},...,\mathbf{LP_{K}})$ and $(\mathbf{UP_{1}},...,\mathbf{UP_{K}})$, we can detect certain faults (especially gradual faults) of the load and unload robot, respectively. However, due to the limited fault information contained in $\mathbf{Y}$, these faults are difficult to predict using sensor data only.

\begin{table}[bt]
\centering
\caption{An example of sensor data. All values are fictional.} \label{tab:sensor}
\begin{tabular}{|l|l|l|l|}
\hline
$\mathbf{Robot}$ &  $\mathbf{Position}$ & $\mathbf{Value}$ & $\mathbf{DateTime}$\\
\hline
Load & $P_{1}$ &  0.05 & 2020-01-01 00:00:00\\
\vdots & \vdots &  \vdots &  \vdots\\
Load & $P_{K}$ &  0.04 & 2020-01-01 00:01:00\\
Unload & $P_{1}$ & 0.04 & 2020-01-01 00:02:00\\
\vdots & \vdots & \vdots& \vdots\\
Unload & $P_{K}$ & 0.04 & 2020-01-01 00:03:00\\
\vdots & \vdots & \vdots & \vdots\\
\hline
\end{tabular}
\end{table}

Therefore, we aim to address the following problem: 
\textit{Suppose there is a complex system $\Delta$ that is composed of several interconnected subsystems $\{\Gamma,\Lambda,...,\Theta\}$. Given a database of logs $\mathbf{X}_{\Delta}$ generated by the entire system $\Delta$ and a database $\mathbf{Y}_{\Theta}$ consisting of time series measured by sensors from a certain subsystem $\Theta$, we assume the set of unique log events in $\mathbf{X}_{\Delta}$ is $\mathbf{C}_{\Delta}=\{C_{1},C_{2},...,C_{N}\}$. Based on $\mathbf{Y}_{\Theta}$, we attempt to select a subset of $\mathbf{C}_{\Delta}$, which is denoted by $\mathbf{C}_{\Theta}=\{C_{1}^{'},C_{2}^{'},...,C_{L}^{'}\}$ with $L\ll N$, resulting in a new database $\mathbf{X}_{\Theta} \subset \mathbf{X}_{\Delta}$ that mainly keeps relevant log events for detecting and predicting faults occurred in the subsystem $\Theta$.} 

Specifically, the system ${\Delta}$ is a lithography machine and the certain subsystem $\Theta$ is the wafer transfer subsystem in this work. 

To solve the above problem, we propose a method consisting of three modules, namely \textit{Log Event Vectorization}, \textit{Selection of Relevant Features} and \textit{Remove of Redundant Features}, each of which will be separately described next. 

\subsection{Log Event Vectorization (Module 1)}
The first module, namely \textit{Log Event Vectorization}, mainly aims at converting  unstructured or semi-structured log events into time series data. Considering the log messages given in Table \ref{tab:event}, it is straightforward to generate a time series for each \textbf{Code} per machine by keeping only the corresponding records.  Without loss of generality, we assume that the smallest unit of time in the original data is seconds. Accordingly, for each log event in a certain machine, we can obtain a time series in the form of $ \{(Value_{t},DateTime_{t})\}_{t \in T}$, meaning that this log event is triggered $Value_{t}$ times at the timestamp $DateTime_{t}$ for $t \in T$, where $T$ represent all time points (an ordered list) when this log event is triggered. 

After some evaluations, we have decided to take
each day as an interval to count the number of times a log event is triggered. Besides, we take the start point of this time interval to represent the time point when this log event is triggered.

\subsection{Selection of Relevant Features (Module 2)}
Given that the load robot and unload robots are very similar, for simplicity, we only consider the load robot when elucidating the proposed method. That is, we assume a multivariate time series database $\mathbf{Y} = (\mathbf{P_{1}},...,\mathbf{P_{K}})$. Note that all these time series are of equal length. Therefore, we assume  $\mathbf{P_{k}} = \{y_{kt}\}_{t \in T}$ for ${k \in \{1,...,K\}}$. Meanwhile, after applying the \textit{Log Event Vectorization} module on the log event database $\mathbf{X}$, we can obtain another multivariate time series database  $\mathbf{Z} = (\mathbf{Z_{1}},...,\mathbf{Z_{n}},...,\mathbf{Z_{N}})$ where $N$ represents the number of unique log features (i.e., log events). Although different log events are usually triggered at different time points, we have taken each day as an interval to count the number of times that each log event is triggered. As a result, for $n \in \{1,2,...,N\}$, $\mathbf{Z_{n}}$ has a fixed length and thus we assume $\mathbf{Z_{n}} = \{z_{ns}\}_{s \in S}$.

\begin{figure}
\includegraphics[width=12.5cm]{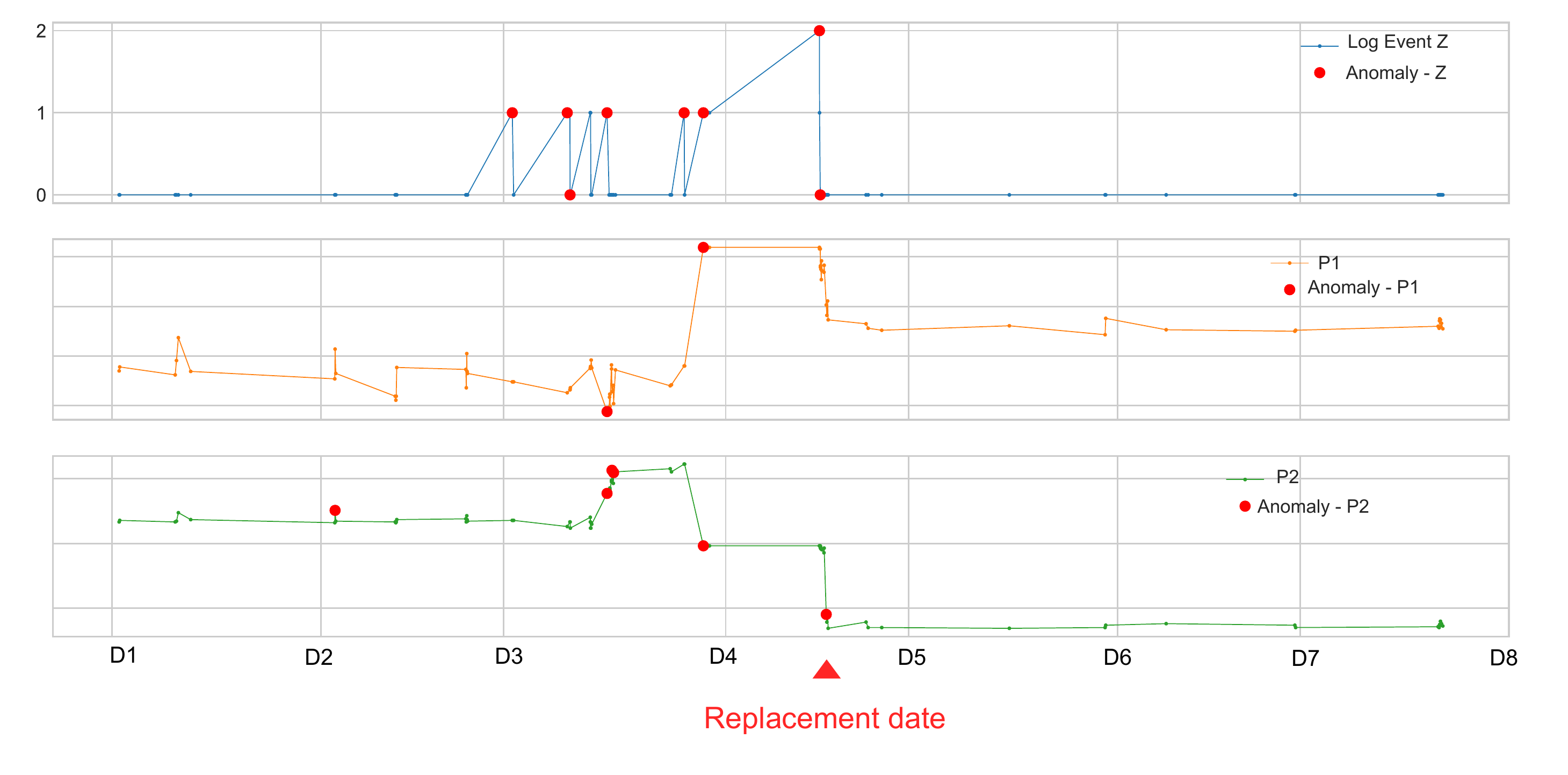}
\caption{An example showing the relevance of a specific log event to detecting and predicting faults of a load robot based on sensor data. The shared x-axis represents the timestamp and the y-axes represent the measured values of each feature. Note that the y-axes of $P_1$ and $P_2$ are intentionally hidden and their timestamps are anonymized.} \label{fig:example}
\end{figure}

To detect faults, for ${k \in \{1,...,K\}}$, we can apply a univariate time series anomaly detector $\phi(\cdot)$ on $\mathbf{P_{k}}$, resulting in $\phi(\mathbf{P_{k}})$. Alternatively, we can apply a multivariate time series anomaly detector $\psi(\cdot)$ on   $\mathbf{Y}$ by jointly considering all $\mathbf{P_{k}}$ for ${k \in \{1,...,K\}}$, leading to $\psi(\mathbf{P_{1}},...,\mathbf{P_{K}})$. As shown in Figure \ref{fig:example} (the two subplots at the bottom), the identification of faults is feasible by applying time series anomaly detectors on $\mathbf{Y}$. However, for gradual faults, just identifying them is not enough. It is also necessary to predict them accurately. Since there is only limited fault information in $\mathbf{Y}$, it is difficult to predict these faults based on $\mathbf{Y}$ alone. Therefore, we attempt to select relevant log features from $\mathbf{Z}$ to better detect and predict faults. For ${k \in \{1,...,K\}}$, by considering $\mathbf{P_{k}}$ as the target variable and $\mathbf{Z_{1}},...,\mathbf{Z_{n}},...,\mathbf{Z_{N}}$ as the prediction variables, it becomes a supervised feature selection problem. Nonetheless, compared to traditional supervised feature selection problems, there are three novel challenges:
\begin{itemize}
    \item The features considered are time series rather than numeric tabular data;
    \item The target and prediction variables are not of equal length, and their timestamps are also different;
    \item Traditional similarity metrics (e.g. Euclidean distance, dynamic time warping) do not give meaningful results when measuring the relevance/similarity of the predictor variable to the target variable.
\end{itemize}

%To make these feature lengths equal, we propose the following interpolation method. Without loss of generality, we take $\mathbf{P_{1}} = \{y_{1t}\}_{t \in T}$ and $\mathbf{Z_{n}} = \{z_{ns}\}_{s \in S}$ as examples. First, we merge the two ordered timestamp lists $T = [t_{1},t_{2},...,t_{A}]$ and $S = [s_{1},s_{2},...,s_{B}]$ into an ordered timestamp list $R = [r_{1},r_{2},...,r_{C}]$. For example, if we have  $T$ = [2020-01-01 00:00:00, 2020-01-01 00:10:00, 2020-01-01 00:20:00] and $S$ = [2020-01-01 00:20:00, 2020-01-01 00:25:00, 2020-01-01 00:30:00], then we will have $R$ = [2020-01-01 00:00:00, 2020-01-01 00:10:00, 2020-01-01 00:20:00, 2020-01-01 00:25:00, 2020-01-01 00:30:00]. As a result, $\mathbf{P_{1}} = \{y_{1r}\}_{r \in R}$ and $\mathbf{Z_{n}} = \{z_{nr}\}_{r \in R}$. For $r \in R$, if $y_{1r}$ does not exist, we fill it with a zero; if $z_{1r}$ does not exist, we fill it with its nearest valid observation.

We now detail why traditional similarity metrics fail to provide meaningful results when trying to find relevant log features. As shown in Figure \ref{fig:example}, we can see that from `D3' some faults started to appear in the robot and became detectable after a certain time based on $\mathbf{P1}$ and $\mathbf{P2}$. These faults disappeared after the replacement of specific components on the `Replacement Date'. Meanwhile, we can observe that the log event $\mathbf{Z}$ was triggered several times before the replacement date. More importantly, it was triggered several times even before the faults became detectable based on $\mathbf{P1}$ and $\mathbf{P2}$. At other times, this log event was not triggered. In other words, the log event $\mathbf{Z}$ can be potentially used to detect and predict these faults. However, if we consider their shapes (wrapped or not) or values (normalised or not), we can see that the log feature $\mathbf{Z}$ is not similar to $\mathbf{P1}$ or $\mathbf{P2}$. To address this problem, we propose a novel similarity metric which consists of three steps as follows.

First, for $k \in \{1,...,K\}$, we apply an appropriate univariate time series anomaly detector $\phi(\cdot)$ on $\mathbf{P_{k}}$, resulting in a time series of anomaly scores  $\mathbf{U_{k}} = \{u_{kt}\}_{t \in T}$. Second, for $n \in \{1,2,...,N\}$, we apply an appropriate univariate time series anomaly detector $\varphi(\cdot)$ on $\mathbf{Z_{n}}$, resulting a time series of anomaly scores  $\mathbf{V_{n}} = \{v_{ns}\}_{s \in S}$. Note that $\phi(\cdot)$ and  $\varphi(\cdot)$ can be different considering that $\mathbf{Z_{n}}$ is sampled at a regular frequency (i.e., an observation per day) but $\mathbf{U_{k}}$ is sampled at an irregular frequency (e.g., multiple observations in day A but no observation in day B). Third, we can select relevant log events by comparing $\mathbf{U_{k}}$ with  $\mathbf{V_{n}}$. Note that the lengths and scales of $\mathbf{U_{k}}$ and $\mathbf{V_{n}}$ may be different, but their peaks should overlap (for detection) or preferably the peak of $\mathbf{V_{n}}$ precedes the corresponding peak of $\mathbf{U_{k}}$ (for prediction) if we compare them using the same timeline. A peak here means a relatively high degree of outlyingness.

\subsubsection{Time series anomaly detector $\phi(\cdot)$}
Since $\mathbf{U_{k}}$ is sampled at an irregular frequency, it may have multiple observations on a given day, but no observations in the following tens of days. However, traditional time series anomaly detection methods usually assume that the input time series is regularly sampled. To circumvent this limitation, we adapt a simple yet effective anomaly detection strategy, which is called \textit{persistence checking}. Specifically, for each time series value $U_{kt}$ in $\mathbf{U_{k}}$, it compares this value with its previous value as the anomaly score, defined as $\phi(U_{kt}) =|U_{kt}-U_{kh}|$ with $t=h+1$.

\subsubsection{Time series anomaly detector $\varphi(\cdot)$} Although $\mathbf{V_{n}}$ is in the form of a time series, we are not concerned about the temporal order of observations when detecting anomalies. Therefore, we can apply a traditional anomaly detector designed for tabular data on it. Specifically, we define $\varphi(V_{ns}) = \frac{|V_{ns}-med(\mathbf{V_{n}})|}{std(\mathbf{V_{n}})}$ as the anomaly score for the sample point $V_{ns}$ in the time series $\mathbf{V_{n}}$, where $med$ and $std$ denote the median and standard deviation of all sample points in $\mathbf{V_{n}}$, respectively. 

\subsubsection{Feature selection via comparing $\phi(\mathbf{U_{k}})$ with $\varphi(\mathbf{V_{n}})$ } We perform feature selection by comparing obtained anomaly scores $\phi(\mathbf{U_{k}})$ and $\varphi(\mathbf{V_{n}})$. Assuming that the robot works continuously for $T$ days, for $\varphi(\mathbf{V_{n}})$, we have an anomaly score per day. However, for $\phi(\mathbf{U_{k}})$, we may have multiple anomaly scores on some days, but no anomaly scores on most days. To make $\phi(\mathbf{U_{k}})$ and $\varphi(\mathbf{V_{n}})$ comparable, we modify $\phi(\mathbf{U_{k}})$ as follows: for each day, if there are  multiple anomaly scores, we take the maximum of these scores as the final anomaly score; if there is no anomaly score, we set the final anomaly score as zero. We denote the modified $\phi(\mathbf{U_{k}})$ as $\hat{\phi}(\mathbf{U_{k}})$, which has the same length as $\varphi(\mathbf{V_{n}})$. Finally, we compute the Kendall's $\tau$ coefficient \cite{kendall1938new} between  $\hat{\phi}(\mathbf{U_{k}})$ and $\varphi(\mathbf{V_{n}})$ for feature selection. A high value of this coefficient indicates the high relevancy of log event $\mathbf{V_{n}}$ to $\mathbf{U_{k}}$ for detecting  faults. By setting a threshold, we can select a subset of log events that are considered to be the most relevant.

\subsection{Removal of Redundant Features (Module 3)}
After selecting a subset of relevant log events in Module 2, we can further reduce the number of selected log events 
by computing the correlation between them. Specifically, we compute the pairwise Kendall's $\tau$ coefficient between selected log events and remove the redundant ones.

\section{Experiments and Preliminary Results}
We use 12 real-world datasets to test our method. A summary of datasets is given in Table \ref{tab:dataset}. Specifically, in module 2, we set a threshold for the similarity coefficient to select 20\% of log events. In module 3, we further remove some highly correlated log features to keep 40 log features.

After constructing the event count matrix based on the selected features, we apply a commonly used unsupervised anomaly detector on it: KNN \cite{ramaswamy2000efficient}. To demonstrate the necessity of feature selection, we also build an event count matrix from all original features and then apply KNN on it. Specifically, a fault is considered detected if the point with the highest anomaly score is on or slightly earlier than the date when the fault is known to happen.

As shown in Table \ref{tab:dataset}, our proposed feature selection method can help improve log-based anomaly detection performance. Specifically, based on the selected log features, KNN was able to accurately detect or predict faults in 11 out of 12 machines. In contrast, KNN can only accurately detect or predict faults in 5 out of 12 machines based on all log features. One possible reason is that the inclusion of many irrelevant log events renders the detection of gradual faults difficult.

\begin{table}
\centering
\caption{Summary of datasets and preliminary experiment results. \textit{Replacement} indicates the date when some components of the robot are replaced (faults always happen on or earlier than this date). \textit{\#Messages} represents the number of log messages. \textit{\#Raw} denotes the number of log features in the original dataset and \textit{\#Selected} denotes the number of selected log features. Besides, AD indicates whether the fault is detected based on the corresponding log features. Note that the value of \textit{Fault} is anonymized, where $GF$ and $SF$ represent gradual fault and sudden fault, respectively.} \label{tab:dataset}
\begin{tabular}{|l|l|l|l|l|l|l|l|l|}
\hline
Machine & Robot &Fault & Replacement & \#Messages & \#Raw(AD) & \#Selected(AD)\\
\hline
1 & Unload &$GF_{1}$ &2021-01-05  & 103294 & 298(No) & 40(Yes)\\
2 & Unload &$GF_{1}$ &2021-01-06 & 90974 & 246(No)  & 40(Yes) \\
4 & Unload &$GF_{1}$ &2021-02-08 & 93729 & 437(No) & 40(Yes) \\
5 & Unload &$GF_{1}$ &2021-02-08 & 76797 & 244(No) & 40(Yes) \\
17 & Unload &$GF_{1}$ &2020-06-03 & 107414  & 217(No) & 40(No) \\
20 & Unload &$GF_{1}$ &2020-06-22 & 106070 & 215(No) & 40(Yes) \\
6 & Load &$SF_{1}$ &2019-06-24 & 86988  & 303(Yes) & 40(Yes) \\
7 & Unload &$SF_{1}$ &2020-03-10 & 95513 & 314(Yes)  &40(Yes)  \\
8 & Load &$SF_{1}$ &2020-07-21 & 23262 & 333(Yes) & 40(Yes) \\
9 & Unload &$SF_{1}$ &2020-07-24 & 34158 & 365(Yes) & 40(Yes)  \\
13 & Load &$SF_{2}$ &2019-12-04 & 94426 & 294(Yes) & 40(Yes) \\
18 & Load &$GF_{2}$ &2020-06-09 & 93875 & 402(No)  & 40(Yes)  \\
\hline
\end{tabular}
\end{table}

\section{Conclusion and Future Work}
In this work we have proposed a simple yet effective feature selection method for log based anomaly detection. This method has been empirically proven to be effective on 12 real-world datasets, but we should note that all twelve datasets were similar in the sense that they were collected from similar machines. In the future, we plan to include more datasets for testing. More importantly, we will try more anomaly detection methods when defining $\phi(\cdot)$ and $\varphi(\cdot)$. On this basis, we can use Granger causality test \cite{arnold2007temporal} to find log events that can be used to predict sensor time series anomalies. Furthermore, we did not specifically consider the causal relationships between different log events. In the future, by constructing a causality graph using the PC-algorithm \cite{spirtes2000causation} on log events, we can investigate the causal relationships between them. As a result, it might be possible to find the root causes of anomalies.

\section{Acknowledgement}
This publication is part of Project 4 of the Digital Twin research programme, a TTW Perspectief programme with project number P18-03 that is (primarily) financed by the Dutch Research Council (NWO).
%
% the environments 'definition', 'lemma', 'proposition', 'corollary',
% 'remark', and 'example' are defined in the LLNCS documentclass as well.
%

\bibliographystyle{splncs04}
\bibliography{bibliography}

\end{document}